\pdfoutput=1

\documentclass[11pt]{article}

\usepackage[final]{acl}
\usepackage{footnote}
\usepackage{times}
\usepackage{latexsym}
\usepackage{diagbox}
\usepackage[T1]{fontenc}

\usepackage[utf8]{inputenc}

\usepackage{microtype}
\usepackage{booktabs}
\usepackage{adjustbox}
\usepackage{amssymb}
\usepackage{enumitem}
\usepackage{xspace}
\usepackage{inconsolata}
\usepackage{xcolor}         

\usepackage{makecell}
\usepackage{amsthm}
\usepackage{bm}
\usepackage{multicol}
\usepackage{colortbl}
\usepackage[most]{tcolorbox}
\usepackage{color}

\definecolor{tabhighlight}{HTML}{e5e5e5}
\usepackage{graphicx}
\usepackage{amsmath}
\usepackage[capitalize]{cleveref}

\usepackage{listings}
\usepackage{color}
\usepackage[most]{tcolorbox}
\usepackage{graphicx} %
\usepackage{booktabs} %
\usepackage{multirow}
\usepackage{xcolor} %
\usepackage{subcaption} %
\usepackage{caption}
\usepackage{float}
\usepackage{colortbl}
\usepackage{tabularx} %
\usepackage{lipsum}

\definecolor{lightblue}{HTML}{ebf3f8}

\definecolor{mediumblue}{HTML}{d7e8f2}

\definecolor{deepblue}{HTML}{c8dfed}

\definecolor{softred}{HTML}{FADBD8}

\definecolor{softorange}{HTML}{FDEBD0}

\definecolor{softpurple}{HTML}{EBDEF0}

\definecolor{pinkpurple}{HTML}{F5EAF7}

\definecolor{coral}{HTML}{F5B7B1}

\usepackage{microtype}
\usepackage{url}
\usepackage{booktabs}
\usepackage{longtable}
\usepackage{placeins}
\definecolor{lightgray}{gray}{0.95}
\definecolor{coral}{rgb}{0.56, 0.93, 0.56}
\definecolor{coral}{rgb}{1.0, 0.5, 0.31}
\definecolor{lavender}{rgb}{0.9, 0.9, 0.98}

\tcbset{
  aibox/.style={
    top=10pt,
    colback=white,
    colframe=black,
    colbacktitle=black,
    enhanced,
    center,
    text width=0.9\linewidth,
    attach boxed title to top left={yshift=-0.1in,xshift=0.15in},
    boxed title style={boxrule=0pt,colframe=white,},
  }
}

\newtcolorbox{AIbox}[2][]{aibox, title=#2,#1}

\title{Condor: Enhance LLM Alignment with Knowledge-Driven \\ Data Synthesis and Refinement} 

\author{Maosong Cao$^{1,*}$, Taolin Zhang$^{1,2,*}$, Mo Li$^{1,2}$, Chuyu Zhang$^{1}$, Yunxin Liu$^{2}$\\ \bf{Haodong Duan$^{1,\dagger}$, Songyang Zhang$^{1,\dagger,\ddagger}$, Kai Chen$^{1,\dagger}$}\\
$^1$Shanghai AI Laboratory $^2$Tsinghua University\\
\texttt{Datasets:\href{https://hf.co/datasets/internlm/Condor-SFT-20K}{https://hf.co/datasets/internlm/Condor-SFT-20K}}\\
\texttt{Github:\href{https://github.com/InternLM/Condor}{https://github.com/InternLM/Condor}}\\
}

\begin{document}
\maketitle

\begin{abstract}

The quality of Supervised Fine-Tuning (SFT) data plays a critical role in enhancing the conversational capabilities of Large Language Models (LLMs).
However, as LLMs become more advanced, 
the availability of high-quality human-annotated SFT data has become a significant bottleneck, 
necessitating a greater reliance on synthetic training data. 
In this work, we introduce \textbf{Condor}, 
a novel two-stage synthetic data generation framework that incorporates  \textbf{World Knowledge Tree} and \textbf{Self-Reflection Refinement} to produce high-quality SFT data at scale. 
Our experimental results demonstrate that a base model fine-tuned on only 20K Condor-generated samples achieves superior performance compared to 
counterparts. 
The additional refinement stage in Condor further enables iterative self-improvement for LLMs at various scales (up to 72B), 
validating the effectiveness of our approach. 
Furthermore, our investigation into the scaling for synthetic data in post-training reveals substantial unexplored potential for performance improvements, 
opening promising avenues for future research.\footnote{This work is done when Taolin Zhang and Chuyu Zhang are on internship at Shanghai AI Laboratory, $*$ means equal contribution, $\dagger$ means corresponding author, $\ddagger$ means project lead.}


\end{abstract}

\section{Introduction}

Large Language Models (LLMs) are experiencing rapid advancements, 
with proprietary models such as like GPT~\citep{achiam2023gpt} and Gemini~\citep{team2023gemini}, 
alongside open-source counterparts such as LLaMA~\citep{dubey2024llama}, Qwen~\citep{yang2024qwen2}, Mistral~\citep{jiang2023mistral}, and Deepseek~\citep{liu2024deepseek}  evolving at an unprecedented pace. 
However, this rapid iteration comes with a significant challenge: 
the depletion of existing high-quality data for Supervised Fine-Tuning (SFT).
Moreover, the internet is increasingly inundated with synthetic data of varying and often questionable quality, 
making it ever more difficult to construct and filter higher-quality training data for LLMs.

 \begin{figure}[t]
  \centering  
\includegraphics[width=0.5\textwidth]{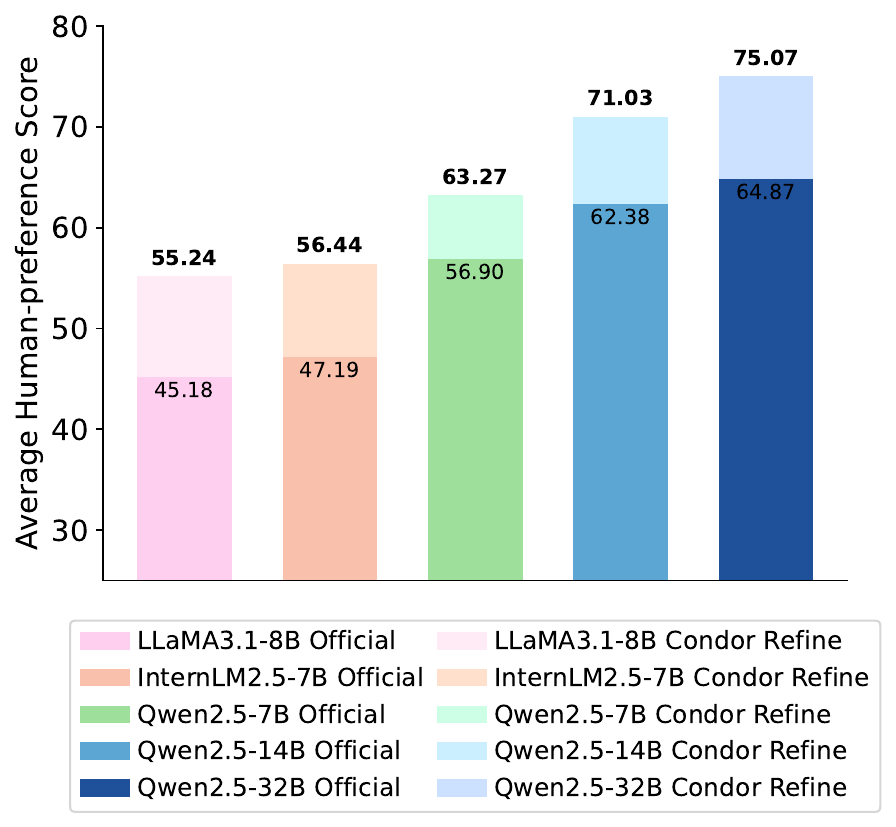}
  \caption{\textbf{Comparison between Official Models and Condor-Refined Models.} The performances are obtained from CompassJudger-1-32B. Different families of popular LLMs benefit greatly from the Condor Refine engine. }
  \label{fig:incremental}
\end{figure}

Empirical studies on SFT training have conclusively shown that both data quality and quantity play crucial roles in enhancing model performance~\citep{shen2024rethinking}. 
While scaling laws suggest that models can achieve extraordinary capabilities when trained on large datasets, 
high-quality data can yield comparable results even at smaller scales. 
This highlights the importance of generating substantial amounts of high-quality data to further enhance large-scale model capabilities.
Recent research introduce various methods for synthesizing SFT training data~\citep{wang2024survey, ding2023enhancing, xu2023wizardlm, yuan2024self, tang2024synthesizing}.
However, these approaches face several limitations: they often rely heavily on existing datasets for seed prompts, lack prompt diversity, and cannot guarantee the quality of the generated data.

To address these challenges, we introduce \textbf{Condor}, 
an automated two-stage pipeline for synthetic data generation 
that effectively leverages existing world knowledge. 
Our approach begins with constructing a comprehensive \textbf{World Knowledge Tree} by prompting LLM to generate tags, and then creating a complete tag chain from coarse to fine through an auto-regressive pipeline. 
After generating diverse questions and original responses using tags, we then employ \textbf{Self-Reflection Refinement} to obtain synthesized SFT data with more higher-quality replies. 
Condor further generates extensive and high-quality SFT data by integrating up-to-date world knowledge tags. 
In this way, Condor not only ensures the relevance and diversity of the content but also maintains scalability in data synthesis.
We conduct experiments on multiple models using the data synthesized by Condor, as shown in Figure~\ref{fig:incremental}, the results indicate that the data synthesized by Condor significantly improves the subjective chat capabilities of the models across the board.


To summarize, our contributions are as follows:

\begin{enumerate}[label={\bf {{$\bullet$}}},,leftmargin=*,topsep=0.5ex,itemsep=-0.5ex,partopsep=0.75ex,parsep=0.75ex,partopsep=0pt,wide,labelindent=0pt]
\item We propose \textbf{Condor Void}, a novel approach incorporating a knowledge inspiration strategy based on the World Knowledge Tree, enabling models to acquire rich world knowledge for generating diverse questions.
\item We further introduce \textbf{Condor Refine}, a self-reflection mechanism that allows models to iteratively refine their responses during the construction phase, resulting in higher-quality responses.
\item We investigate the scaling and self-iteration techniques for Condor datasets, offering valuable insights and potential pathways for the scalable expansion of synthetic data.
\end{enumerate}

\section{Related Work}

\subsection{Synthesizing Instruction Tuning Data}
With the rapid advancement of Large Language Models (LLMs)~\citep{achiam2023gpt, team2023gemini, dubey2024llama, yang2024qwen2, jiang2023mistral, liu2024deepseek}
and Large Multi-Modality Models (LMMs)~\citep{radford2021learning, zhu20233d, zhang2024vision, zhang2024boostadapter, liu2024visual,  chen2024internvl}, 
the demand for high-quality training data continues to grow. 
As a result,  synthetic data generation has become increasingly crucial for model development. 
While high-quality synthetic data can significantly enhance model performance, low-quality synthetic data may lead to model degradation or collapse.


WizardLM~\citep{xu2023wizardlm} introduces an Evol-Instruct scheme,
which expands an initial question dataset by generating new questions through techniques such as deep thinking, adding constraints, and reconstruction. 
However, the quality and diversity of the generated data are strongly influenced by the characteristics of the original dataset.
Similarly, Self-Reward~\citep{yuan2024self} builds on a seed dataset, 
enabling the model to generate questions via few-shot prompting and produce multiple corresponding responses.
These responses are subsequently evaluated by a reward model and labeled as training data for reinforcement learning with human feedback (RLHF).

Magpie\citep{xu2024magpie} explores leveraging models themselves to generate synthetic data by utilizing chat templates as prompts to directly produce dialogues. 
However, this approach is not universally applicable to all model types, and therefore lacks generalization ability.
Additionally, the synthetic data generated often lacks stability, 
containing significant noise that requiring extensive filtering to maintain quality. 
Agent-based data synthesis methods, 
such as MATRIX-Gen~\citep{tang2024synthesizing}, 
simulate multiple world scenarios using a multi-agent framework and collect dialogue data from interactions within these scenarios. 
While effective in some cases, this method is time-consuming, labor-intensive, and limited by the need for multiple models to operate concurrently.

In contrast to these methods, Condor eliminates the dependence on a seed dataset, 
requiring only tags as seeds for expansion and generation. 
This enables exceptional diversity and scalability in data production. 
Furthermore, Condor handles all data generation tasks with a single model,
significantly reducing costs while maintaining high efficiency. 




 \begin{figure*}[ht]
  \centering  
\includegraphics[width=\textwidth]{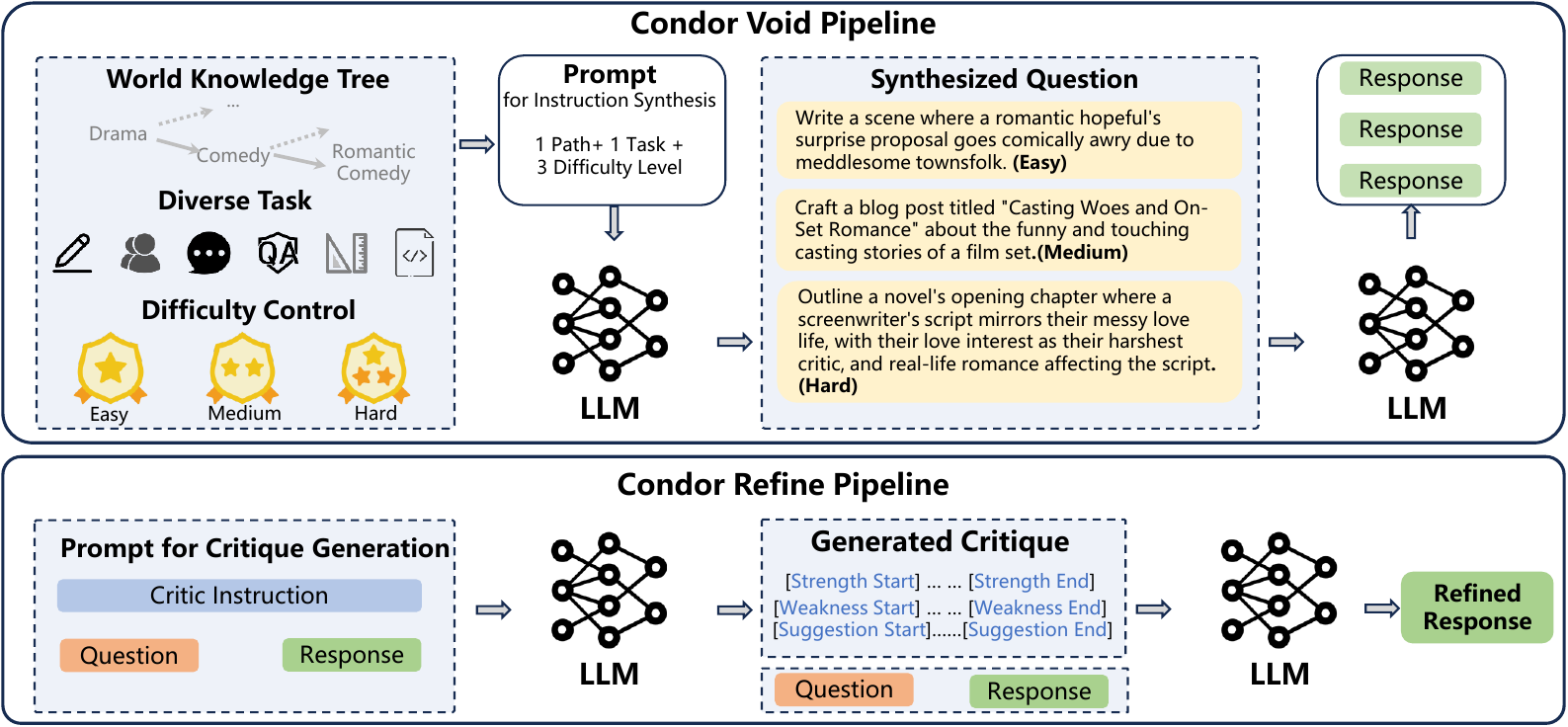}
\vspace{-2mm}
  \caption{\textbf{The Pipeline of Condor Engine.} The pipeline is mainly divided into (1) Data synthesis: LLM is inspired from the \textbf{World Knowledge Tree} with different task diversity and difficulty to construct the original question-answer dataset $\boldsymbol{{D}_{V}}$;
  (2) Data refinement: LLM is utilized to reflect on the synthesized data and generate critiques for refinement and improve the quality of responses, producing the dataset $\boldsymbol{{D}_{R}}$. }
  \label{fig:overall2}
\end{figure*}

\subsection{LLM Self-Iteration}

Recent studies have explored methods for large language models (LLMs) to improve themselves through iterative self-enhancement. 
For instance, I-SHEEP~\citep{liang2024isheepselfalignmentllmscratch} and Self-Instruct~\citep{wang2023selfinstructaligninglanguagemodels} utilize seed data and LLMs to generate new instructions. 
However, these methods often produce instructions that closely resemble those in the seed dataset, limiting the diversity and novelty of the generated data. 
\cite{tao2024surveyselfevolutionlargelanguage} categorizes existing self-iteration methods,
highlighting works such as 
self-instruct~\citep{wang2023selfinstructaligninglanguagemodels}, 
self-improving~\citep{huang2022largelanguagemodelsselfimprove},
and self-training~\cite{gulcehre2023reinforcedselftrainingrestlanguage}. Specifically, the self-improving framework proposed by 
\cite{huang2022largelanguagemodelsselfimprove} 
enhances the reasoning abilities of LLMs without requiring labeled data 
by generating multiple reasoning paths and answers, 
then selecting high-confidence answers through a self-consistency mechanism. 


Reinforced Self-Training (ReST) \cite{gulcehre2023reinforcedselftrainingrestlanguage} 
achieves self-iteration by generating multiple outputs, scoring them with a reward model, and fine-tuning the LLM using the highest-scoring samples. 
While effective, this approach depends on an external reward model to provide additional knowledge. 
Other notable works, such as 
Reflexion~\citep{shinn2023reflexionlanguageagentsverbal}, 
Self-Refine~\citep{madaan2023selfrefineiterativerefinementselffeedback}, 
and Refiner~\citep{paul2024refinerreasoningfeedbackintermediate},
propose self-iteration techniques where models iteratively refine their answers based on feedback from specific tasks. 
However, these methods are primarily designed for specific domains like mathematics, limiting their ability for generalization.


By contrast, 
Condor achieves self-improvement on a wide range of human-preference datasets solely using the model’s own capabilities, enabling broader applicability while maintaining scalability and efficiency.

\section{Methodology}
\subsection{Pipeline Overview}

As shown in \Cref{fig:overall2}, the pipeline of Condor is divided into two stages: 
\textbf{data synthesis} and \textbf{data refinement}. 
In the data synthesis stage, we begin by introducing the \textbf{World Knowledge Tree}, which serves as a foundation of tags for data generation. Next, we apply task and difficulty expansion to enhance the diversity and complexity of questions under each tag, leading to the creation of the initial synthetic QA dataset $\boldsymbol{{D}_{V}}$.
In the data refinement stage, we employ a \textbf{Self-Reflection Refinement} strategy, 
enabling the model to iteratively optimize the generated responses and yield $\boldsymbol{{D}_{R}}$. 
Notably, we utilize a single model during the entire pipeline for better efficiency. 


\subsection{Condor Void}
In the data synthesis stage, we generate the synthetic data based on existing world knowledge. This stage consists of two key components: \textbf{(1) World Knowledge Tree}  and \textbf{(2) Q\&A pair generation}. The dataset produced from this stage is referred to as $\boldsymbol{{D}_{V}}$.

\subsubsection{World Knowledge Tree}

To construct a comprehensive multi-level tag system that encapsulates diverse aspects of world knowledge, 
we propose a methodology that enhances the model’s capacity for inspiration and knowledge stimulation. 
We define a set of root tags, \textit{i.e.}, \( R = \{ r_1, r_2, \ldots, r_n \} \), 
representing broad themes such as \textit{marriage}, \textit{entertainment}, and \textit{artificial intelligence}. 
Using large language models (LLMs), we expand each root tag into a set of detailed leaf tags, 
denoted as \( L_i = \{ l_{i1}, l_{i2}, \ldots, l_{im} \} \). 
This hierarchical structure forms the foundation of the World Knowledge Tree, \(\mathcal{T}\), 
which can be expressed as a union of subtrees rooted at each \( r_i \): \(\mathcal{T} = \bigcup_{i=1}^{n} \{ (r_i, L_i) \}\). 

To ensure the tree remains reflective of real-world themes and contexts,
we augment it with additional tags collected from trending topics on platforms such as Zhihu and Reddit, 
denoted as \( S = \{ s_1, s_2, \ldots, s_k \} \). 
This creates an enriched tree, \(\mathcal{T} = \bigcup_{i=1}^{n} \{ (r_i, L_i \cup S_i) \}\), 
where \( S_i \subseteq S \) ensures each root branch is contextually informed and relevant.
Moreover, the World Knowledge Tree is designed to be dynamic, 
adapting over time with an update function \( U(\mathcal{T}, t) \) that  that continuously integrates new information.
This robust, evolving framework, comprising over 8,400 tags, is pivotal for maintaining the current and applicable nature of knowledge required for the iteration of LLMs.

\subsubsection{Q\&A Pair Generation}

After acquiring the seed tags, 
we employ various methods to enhance the diversity of the generated questions. 
First, we identify and summarize the most common chat scenarios in human interactions with LLMs, 
such as daily conversation, creative tasks, and role-playing (the total 7 tasks with detailed descriptions are provided in Figure~\ref{taskdef}). 

The model is then required to deeply engage in a specific task under the corresponding tag theme to generate questions. This approach enhances the diversity of the generated questions and encourages knowledge extrapolation, allowing the model to explore themes it may not have frequently encountered—or not encountered at all—in a particular scenario.
To further expand the range of questions, we implement difficulty control, 
prompting the model to adapt to three difficulty levels and generate questions of varying complexity.
This step ensures greater diversity and richness in the generated data, contributing to a more well-rounded and comprehensive dataset.
Finally, we obtain Condor Void datasets with approximately 200k Q\&A pairs, \textit{i.e.}, $|\boldsymbol{{D}_{V}}|\approx200,000$.


\subsection{Condor Refine}
In this stage, we enable the model to engage in self-reflection and critical examination of its own responses, 
followed by revisiting the questions to generate higher-quality replies. 
The resulting refined dataset is referred as $\boldsymbol{{D}_{R}}$.

The initial QA pairs contain responses directly generated by the model, which are inherently limited by the model's current capabilities. Therefore, we implement data refinement to further improve the quality of the responses.
We carefully develop a set of fixed templates that guide the model in identifying both strengths and weaknesses in its responses, 
leading to specific suggestions for improvement.
Subsequently, we utilize these self-generated improvement suggestions to prompt the model to produce enhanced responses. 
The model is tasked with preserving the effective elements of its original responses while addressing identified weaknesses to generate superior responses.
Complete examples of QA pairs that have undergone the entire Condor pipeline are provided in the Appendix.
Finally, we obtain 200k high-quality refinement datasets from $\boldsymbol{{D}_{V}}$, \textit{i.e.}, $|\boldsymbol{{D}_{R}}|\approx200,000$.

\section{Experiments and Results}

\begin{table*}[!t]
    \centering
    \small
    
\resizebox{\textwidth}{!}{
    \begin{tabular}{lcccccc}
    \toprule
        {\multirow{2}{*}{Datasets}} & {\multirow{2}{*}{Score Range}} & {Qwen2.5-7B-Instruct} & \multicolumn{2}{c}{Qwen2.5-7B-Base} & \multicolumn{2}{c}{Qwen2.5-7B-Instruct} \\
        \cmidrule(lr){3-3} \cmidrule(lr){4-5} \cmidrule(lr){6-7}
         & &  Official & Condor Void & Condor Refine & Condor Void & Condor Refine \\
    \midrule
        CompassArena~\citep{2023opencompass} & 0\textasciitilde100 & 33.8 & 32.43 & 36.13 & 35.17 & \textbf{40.12}  \\ 
        FoFo~\citep{xia2024fofo} & 0\textasciitilde1 & \textbf{0.52} & 0.41 & 0.45 & 0.50 & 0.47 \\ 
        AlignBenchv1.1~\citep{liu2023alignbench} & 0\textasciitilde10 & 6.22 & 6.04 & 6.20 & \textbf{6.24} & 6.20  \\
        AlpacaEvalv2~\citep{dubois2024length} & 0\textasciitilde10 &34.66 & 32.84 & 44.10 & 38.76 & \textbf{44.60}   \\ 
        ArenaHard~\citep{li2024crowdsourced} & 0\textasciitilde100 & 53.65 & 42.44 & 60.87 & 53.56 & \textbf{61.53}  \\
        FollowBench~\citep{jiang2023followbench} &0\textasciitilde1 & 0.84 & 0.85 & 0.83 & \textbf{0.88} & 0.85 \\
        MTBench101~\citep{bai2024mt} & 0\textasciitilde10 & \textbf{8.60} & 8.22 & 8.37 & 8.41 & 8.43  \\
        WildBench~\citep{lin2024wildbench} &-100\textasciitilde100 & 16.71 & 14.09 & 30.13 & 21.54 & \textbf{32.70} \\ 
            \rowcolor{tabhighlight} \textbf{Average} & 0\textasciitilde100 & 58.02 & 54.09 & 60.03 & 59.14 & \textbf{61.29} \\

    \bottomrule
    \end{tabular}
}
    \vspace{-2mm}
    \caption{\textbf{Performance Comparison on Human-preference Benchmarks Judged by GPT4o-0806.} The results demonstrate that Condor Refine significantly enhances the subjective chat capability of models. The \textbf{Average Score} is calculated by normalizing the scores for each dataset to a percentage scale and then taking their average.}
    \label{tab:main ablation 2}
\end{table*}

\begin{table*}[!t]
    \centering
    \small
\resizebox{\textwidth}{!}{
    \begin{tabular}{lccccc}
    \toprule
        {\multirow{2}{*}{Datasets}}& {Qwen2.5-7B-Instruct} & \multicolumn{2}{c}{Qwen2.5-7B-Base} & \multicolumn{2}{c}{Qwen2.5-7B-Instruct} \\
        \cmidrule(lr){2-2} \cmidrule(lr){3-4} \cmidrule(lr){5-6}
         & Official & Condor Void & Condor Refine & Condor Void & Condor Refine \\
    \midrule
    BoolQ~\citep{clark2019boolqexploringsurprisingdifficulty} & 86.06 & \textbf{87.83} & 86.82 & 87.34 & 86.33 \\
        GPQA Diamond~\citep{rein2023gpqagraduatelevelgoogleproofqa} & 35.35 & 34.34 & \textbf{35.86} & 35.35 & 35.35 \\
        Math~\citep{hendrycks2021measuringmathematicalproblemsolving} & \textbf{83.21} & 82.31 & 81.91 & 82.67 & 82.47 \\
        GSM8K~\citep{cobbe2021trainingverifierssolvemath} & \textbf{92.27} & 91.21 & 90.75 & 91.43 & 91.96 \\
        GaokaoBench~\citep{zhang2024evaluatingperformancelargelanguage} & 60.42 & \textbf{63.87} & 63.64 & 60.69 & 60.13 \\
        MMLU-Pro~\citep{wang2024mmluprorobustchallengingmultitask} & 54.70 & 52.15 & 50.70 & \textbf{54.73} & 53.02 \\
        DROP~\citep{dua2019dropreadingcomprehensionbenchmark} & 80.81 & 79.31 & 78.50 & \textbf{81.53} & 80.46 \\
        IFEval loose~\citep{zhou2023instructionfollowingevaluationlargelanguage} & 81.70 & 81.70 & 79.85 & \textbf{83.36} & 80.96 \\
        HumanEval~\citep{chen2021evaluating} & \textbf{89.63} & 89.02 & 85.98 & 87.20 & 86.59 \\
        Race-high~\citep{lai-etal-2017-race} & 85.05 & 84.88 & 83.56 & \textbf{85.16} & 84.28 \\
        MBPP~\citep{austin2021program} & 74.32 & 74.32 & 73.54 & 74.32 & \textbf{74.71}\\
        CMO~\citep{2023opencompass} & 22.60 & 25.00 & \textbf{28.85} & 23.08 & 20.19 \\
        \rowcolor{tabhighlight} \textbf{Average} & 70.51 & 70.50 & 70.00 & \textbf{70.57} & 69.70 \\

    \bottomrule
    \end{tabular}
}
    \vspace{-2mm}
    \caption{\textbf{Performance Comparison on Knowledge-Based Benchmarks.} The score range across all the tasks are 0\textasciitilde100. 
    When trained on high-quality SFT data, 
    the differences in the model's performance on knowledge-intensive question-answering tasks are minimal. }
    \label{tab:obj ablation}
    \vspace{-2mm}
\end{table*}

\subsection{Training and Evaluation Settings}
\paragraph{Training Settings.}
We use xTuner~\citep{2023xtuner} as the training framework. To ensure fairness in comparison, we set the initial learning rate to 2e-5 and train for 3 epochs in all experiments. Unless specifically stated, we start training from the base model and compare it with the official RLHF model.

\paragraph{Evaluation Settings.}
We use OpenCompass~\citep{2023opencompass} and employ greedy inference to uniformly evaluate all models, ensuring fairness across various datasets. We selected 8 human-preference benchmarks for evaluation the chat capabilities of the models. 
We evaluate the results and reports the average normalized score at the percentage scale. 
Additionally, due to the high cost of conducting subjective evaluations with a paid API model, 
we use GPT4o as the Judge Model for main results only, 
and we judge with the open-source CompassJudger-1-32B~\citep{cao2024compassjudger} in ablation study and scaling experiments.
Furthermore, we also select a range of knowledge-based Q\&A datasets for groundtruth-based evaluations. 

\subsection{Data Synthesis Settings}
Only one model is required for data synthesis and data refinement throughout the entire pipeline. In the main experiments, we use 200k data synthesized by Qwen2.5-72B for model training. Additionally, for SOTA comparisons and self-interaction experiments, we also generate 200k data points using Qwen2.5-7B to test the effectiveness of the Condor pipeline.

\subsection{Main Results}

Using Condor data generated by Qwen2.5-72B-Instruct, 
we first train both the Base and Instruct versions of Qwen2.5-7B and 
compare their performance against the official model on both Chat and Knowledge benchmarks.

\begin{figure}[!t]
  \centering  
\includegraphics[width=0.45\textwidth]{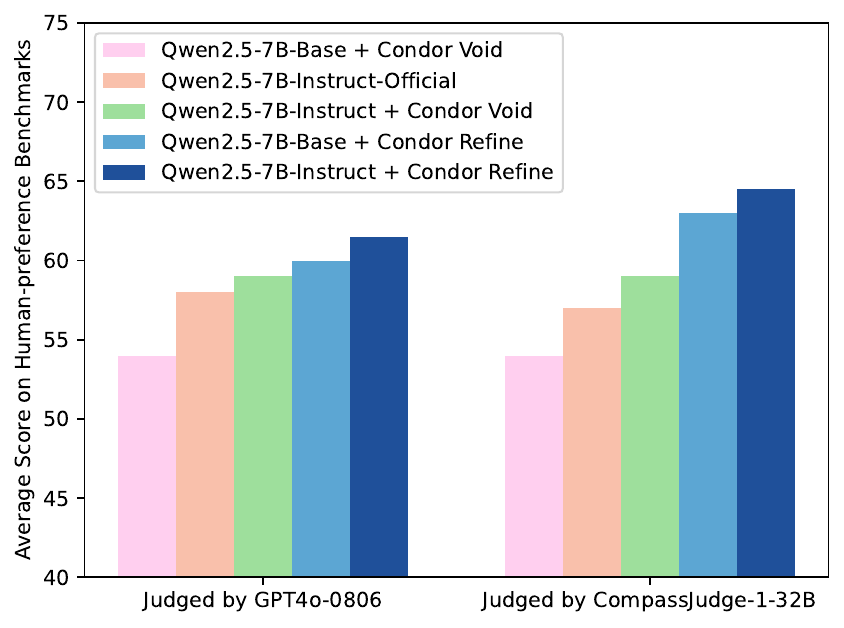}
  \vspace{-2mm}
  \caption{\textbf{Average Score on Human-preference Benchmarks.} 
  CompassJudger-1-32B and GPT-4o provide a consistent trend in scoring.}
  \label{fig:ablation}
  \vspace{-2mm}
\end{figure}

\noindent
\textbf{Human-preference Evaluation. }
We select several widely-used human-preference benchmarks and calculate the final normalized average scores on these datasets (detailed calculation logic is provided in Appendix\ref{scoreappendix}).
We use GPT-4o as the Judge Model, 
and the average results are reported in \Cref{tab:main ablation 2}.
As shown in the table, the results of training the Base model with $\boldsymbol{{D}_{R}}$ surpassed the official model on almost all human-perference benchmarks. 
It is noteworthy that this stage only employs SFT and does not include RLHF as applied in the official model, 
highlighting the significant potential of Condor's constructed data in enhancing the model's subjective chat capabilities. 
Furthermore, continue tuning on the Instruct model further improves performance, 
demonstrating that the data synthesized by Condor complements the original model's capabilities and can enhance human-preference performance without introducing conflicts.
To further verify these results, 
we also conducted evaluations using CompassJudger-1-32B~\citep{cao2024compassjudger}. 
As shown in \Cref{fig:ablation}, 
evaluations using both Judge Models reached a consistent conclusion.


\noindent
\textbf{Knowledge Performance Evaluation. }
We further conduct experiments on several knowledge-based benchmarks to assess the model's objective question-answering capabilities. 
The detailed results are presented in \Cref{tab:obj ablation}. 
It is worth noting that our synthetic data doesn't specifically include knowledge-based QAs like the official models, 
yet it achieves comparable or even superior results. 
The outcome indicates that our dataset not only enhances the model's human-preference performance but also preserves its objective knowledge QA capabilities.
Additionally, the results demonstrate that the model's chat ability and knowledge capability are largely independent. 
The knowledge capability primarily relies on the foundational base model's abilities, 
whereas the chat ability can be independently improved through SFT and RLHF.

\begin{table}[!t]
    \centering
    \adjustbox{width=0.5\textwidth}{
        \begin{tabular}{@{\extracolsep{\fill}}lccc}
            \toprule
            \textbf{Method} & \textbf{Avg Sub Score} & \textbf{Data Size} & \textbf{Synth Model} \\
            \midrule
            Tulu v2~\citep{ivison2023camels} & 24.65 & 326k & Mixed  \\
            Evol Instruct~\citep{xu2023wizardlm} & 34.33 & 70k & GPT  \\
            
            WildChat~\citep{zhao2024wildchat} & 39.53 & 520k & GPT Mixed  \\
            Magpie~\citep{xu2024magpie} & 55.67 & 1M & Qwen2.5-72B  \\
            Qwen2.5-7B-Instruct & 58.02 & 500k+ &  Unknown \\
            Condor 7B & 58.38 & 200k & Qwen2.5-7B \\
            Condor 72B & \textbf{60.03} & 200k & Qwen2.5-72B \\\bottomrule
        \end{tabular}
    }
    \vspace{-2mm}
    \caption{\textbf{Comparison Results with Other Data Synthesis Methods.}
    We adopt SFT data generated by various approaches to tune Qwen2.5-7B-Base and use GPT-4o as the judge model.}
    \label{tab:sota}
    \vspace{-2mm}
\end{table}

\noindent
\textbf{Comparing with the State-of-the-art. }
We generate two sets of Condor data using Qwen2.5-7B and Qwen2.5-72B 
and compare them with existing state-of-the-art synthetic datasets. 
We use the data obtained from various data synthesis methods to train starting from the Qwen2.5-7B-Base, and evaluate the model with GPT-4o judge. 
As shown in \Cref{tab:sota}, 
Condor data generated by both the 7B and 72B models significantly outperforms other data synthesis methods and even surpasses the official model after RLHF.
This observation indicates that self-improvement is achievable even at the 7B scale, which will be further analyzed in detail in Section~\ref{self iter}. Moreover, Condor requires only a single model for the entire data synthesis and refinement pipeline with and mach smaller volume of data than most synthetic datasets, highlighting its efficiency and effectiveness.


\subsection{Ablation Study}

We further investigate the impact of model type, size, and question difficulty on Condor training through a series of ablation experiments using Condor data generated by Qwen2.5-72B-Instruct. 
For the type ablation, 
we evaluate models of different families, 
including Qwen, InternLM, and Llama. 
For the size ablation, 
we test Qwen2.5 at 7B, 14B, and 32B parameter scales. 
Additionally, in the difficulty ablation,
we train models with datasets containing questions of varying difficulty levels.
In all experiments, we utilize CompassJudger as judger for subjective evaluation.


\begin{table*}[!htbp]
    \centering
    \small
    \resizebox{0.9\textwidth}{!}{
    \begin{tabular}{lcccccc}
    \toprule
        {\multirow{2}{*}{Datasets}}& \multicolumn{2}{c}{Qwen2.5-7B} & \multicolumn{2}{c}{Internlm2.5-7B} & \multicolumn{2}{c}{Llama3.1-8B} \\
        \cmidrule(lr){2-3} \cmidrule(lr){4-5} \cmidrule(lr){6-7}
         & Official & Condor Refine & Official & Condor Refine & Official & Condor Refine \\
    \midrule
        CompassArena & 34.95 & \textbf{43.62} & 34.03 & \textbf{42.92} & 9.23 & \textbf{20.48} \\
        FoFo & 0.45 & \textbf{0.48} & 0.40 & \textbf{0.41} & 0.37 & \textbf{0.50} \\
        AlignBenchv1.1 & 6.16 & \textbf{6.26} & 5.64 & \textbf{5.79} & 4.66 & \textbf{5.29} \\
        AlpacaEvalv2 & 32.42 & \textbf{60.00} & 26.34 & \textbf{55.53} & 24.1 & \textbf{47.33} \\
        ArenaHard & 54.72 & \textbf{61.37} & 16.19 & \textbf{36.02} & 31.33 & \textbf{45.73} \\
        FollowBench & \textbf{0.86} & 0.84 & \textbf{0.81} & 0.79 & 0.83 & \textbf{0.84} \\
        MTBench101  & \textbf{8.37} & 8.18 & 8.00 & \textbf{8.07} & 8.18 & \textbf{8.19} \\
        WildBench & 15.69 & \textbf{31.13} & -13.37 & \textbf{17.71} & -2.24 & \textbf{20.45}   \\
       \rowcolor{tabhighlight}\textbf{Average} & 56.90 & \textbf{63.27} & 47.19 & \textbf{56.44} & 45.18 & \textbf{55.24} \\
       \bottomrule 
    \end{tabular}
    }
    \caption{\textbf{Performance Comparison across Different Types of Models.} Condor Refine shows strong adaptability over different types of models.}
    \label{tab:type}
\end{table*}

\begin{table*}[!ht]
    \centering
    \small
    \resizebox{0.9\textwidth}{!}{
    \begin{tabular}{lcccccc}
    \toprule
        {\multirow{2}{*}{Datasets}}& \multicolumn{2}{c}{Qwen2.5-7B} & \multicolumn{2}{c}{Qwen2.5-14B} & \multicolumn{2}{c}{Qwen2.5-32B} \\
        \cmidrule(lr){2-3} \cmidrule(lr){4-5} \cmidrule(lr){6-7}
         & Official & Condor Refine & Official & Condor Refine & Official & Condor Refine \\
    \midrule
        CompassArena & 34.95 & \textbf{43.62} & 35.78 & \textbf{55.05} & 40.18 & \textbf{60.18} \\
        FoFo & 0.45 & \textbf{0.48} & 0.55 & \textbf{0.58} & 0.59 & \textbf{0.63} \\
        AlignBenchv1.1 & 6.16 & \textbf{6.26} & 6.56 & \textbf{6.66} & 6.78 & \textbf{6.88} \\
        AlpacaEvalv2 & 32.42 & \textbf{60.00} & 33.66 & \textbf{71.18} & 35.53 & \textbf{74.41} \\
        ArenaHard & 54.72 & \textbf{61.37} & 71.00 & \textbf{74.95} & 76.82 & \textbf{86.83} \\
        FollowBench& \textbf{0.86} & 0.84 & \textbf{0.92} & 0.90 & \textbf{0.94} & 0.92\\
        MTBench101 & \textbf{8.37} & 8.18 & \textbf{8.43} & 8.30 & \textbf{8.49} & 8.33 \\
        WildBench & 15.69 & \textbf{31.13} & 23.42 & \textbf{40.80} & 22.46 & \textbf{45.84} \\
        \rowcolor{tabhighlight}\textbf{Average} & 56.90 & \textbf{63.27} & 62.38 & \textbf{71.03} & 64.87 & \textbf{75.07} \\

    \bottomrule
    \end{tabular}
    }
    \caption{\textbf{Performance Comparison across Different Model Sizes.} Qwen2.5-7B, 14B, and 32B all benefit from Condor Refine by a significant margin.}
    \label{tab:size}
\end{table*}

\noindent
\textbf{Model Type. }
From \Cref{tab:type}, it is evident that training on $\boldsymbol{{D}_{R}}$ consistently improves performance across different models. 
Almost all models demonstrate significant enhancements on all subjective evaluation datasets. 
Compared to the official model, Qwen2.5-7B shows an improvement more than $6\%$ ($56.9\%$ to $63.3\%$) after training on Condor Refine.
InternLM\footnote{We apply the SFT to an internal version of InternLM2.5-7B, which is enhanced with instruction data focused on mathematics and coding.} and Llama exhibit an even larger improvement by about $10\%$. 
These results highlight that the quality of the base models plays a crucial role in determining the conversational performance of the final Instruct model.

\noindent
\textbf{Model Size. }
We further evaluate the adaptability of models of different sizes to $\boldsymbol{{D}_{R}}$. 
\Cref{tab:size} show results aligning with the observations from \Cref{tab:type}. 
Interestingly, we find that the improvement in performance for the 14B and 32B models is notably greater compared to the 7B variant. 
This suggests that  
larger models demonstrate a greater capacity to learn and benefit from the refined data and achieve better performance.

 \begin{figure}[t]
  \centering  
  \includegraphics[width=0.48\textwidth]{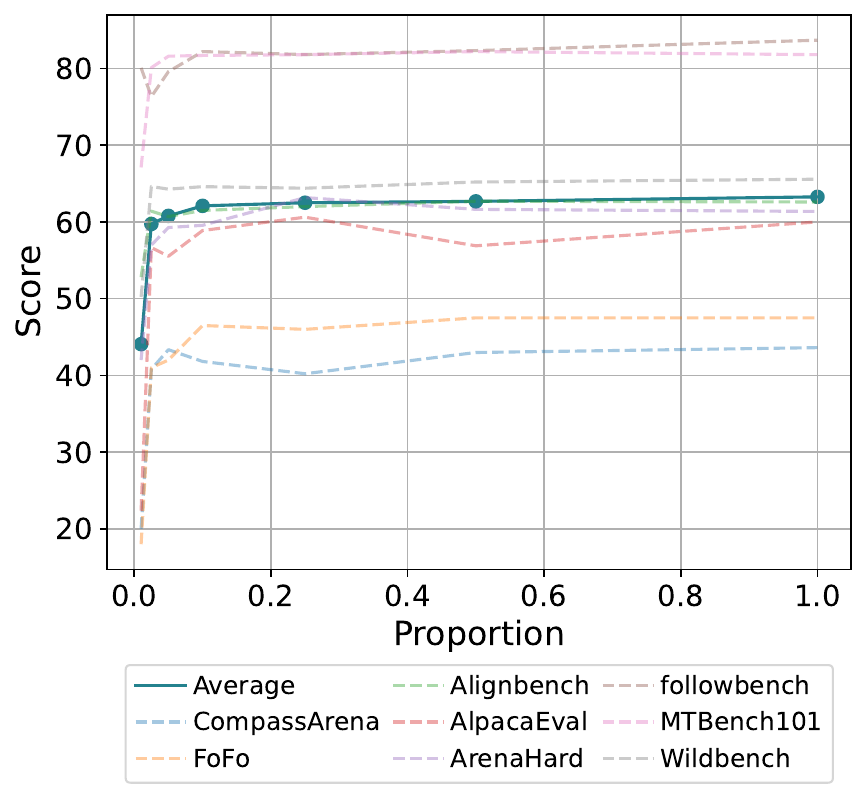}
  \vspace{-7mm}
  \caption{\textbf{Performance Comparison under Different Proportions of Training Data.} "Average" represents the average score across 8 benchmarks, 
  and the results show a clear and consistent improvement as the size of the training data increases (Detailed scores in \Cref{tab:totalnumber}). }
  \label{fig:ablation_radar}
  \vspace{-3mm}
\end{figure}

\noindent
\paragraph{Task Difficulty.}
\Cref{tab:level} presents the performance of models trained with Condor data of 
varying difficulty levels.
The results show that tasks of increasing difficulty contribute to greater performance improvements, 
with more challenging tasks yielding higher average scores. 
Additionally, combining all three difficulty levels during training further enhances performance, resulting in an average improvement of 0.34\%.

\begin{table}[!t]
    \centering
    \adjustbox{width=0.5\textwidth}{
        \begin{tabular}{@{\extracolsep{\fill}}lcccc}
            \toprule
            \textbf{Different level} & \textbf{Exp. Easy} & \textbf{Exp. Medium} & \textbf{Exp. Hard} & \textbf{Exp. All} \\ \midrule
        Easy Data& \checkmark &  &  & \checkmark \\ 
        Medium Data&  & \checkmark &  & \checkmark \\ 
        Hard Data&  &  & \checkmark & \checkmark \\ \midrule
            CompassArena & 42.22 & 38.95 & 43.07 & \textbf{43.62} \\
            FoFo & 0.45 & 0.46 & 0.47 & \textbf{0.48} \\
            AlignBenchv1.1 & 6.18 & 6.22 & 6.16 & \textbf{6.26} \\
            AlpacaEvalv2 & 58.76 & 56.89 & 58.51 & \textbf{60.00} \\
            ArenaHard & 58.55 & 61.02 & \textbf{61.81} & 61.37 \\
            FollowBench & 0.83 & \textbf{0.86} & 0.84 & 0.84 \\
            MTBench101 & \textbf{8.18} & 8.17 & 8.16 & \textbf{8.18} \\
            WildBench & 26.98 & 29.41 & \textbf{31.53} & 31.13 \\ 
            \rowcolor{tabhighlight} \textbf{Average Score} & 61.71 & 62.14 & 62.93 & \textbf{63.27} \\
            \bottomrule
        \end{tabular}
    }
    \caption{\textbf{Ablations on Different Levels of Task Difficulty.} Harder tasks yield greater improvements.}
    \label{tab:level}
\end{table}
\subsection{Scaling of Condor Data }
The performance of models within the Condor pipeline improves with respect to the increasing amount of training data, tags, and tasks.
To investigate these relationships in detail, we conduct comprehensive experiments, as outlined below.

\noindent
\textbf{Training Data Amount. }
We visualize the relative performance under different sampling proportions of training data in \Cref{fig:ablation_radar}.
Surprisingly, we find that the model retains a high percentage of its performance even when trained on only a small proportion of the data. 
Additionally, while some performance fluctuations are observed on specific benchmarks due to the relatively smaller data sizes -- the overall average score consistently improves as the total amount of training data scales up.

\begin{figure}[t]
  \centering  
\includegraphics[width=0.45\textwidth]{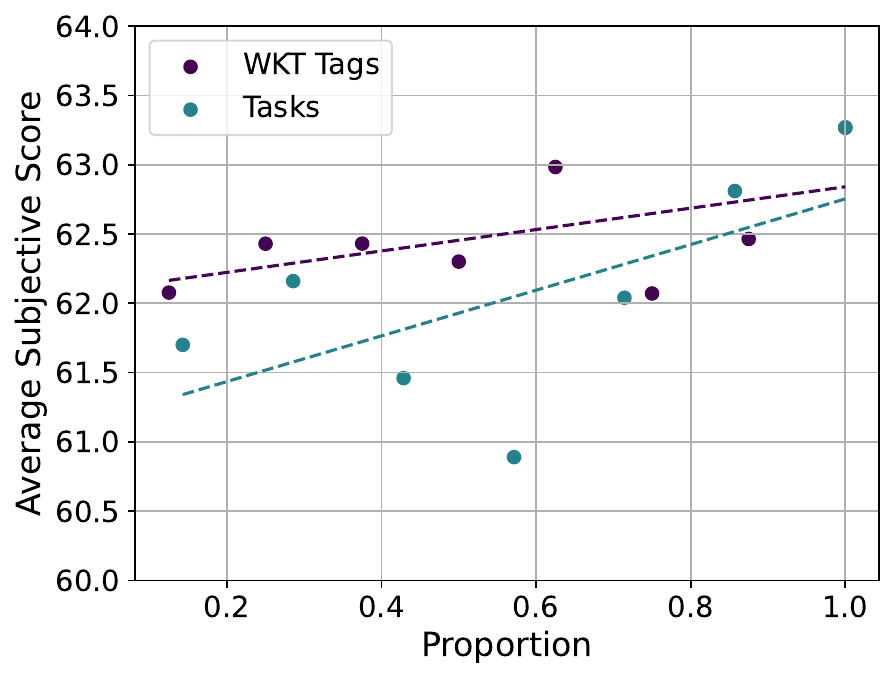}
  \caption{\textbf{Performance Results under Different Proportion of WKT Tags and defined Tasks.} Tasks have a more significant influence than tags, as evidenced by the significant drop with a small proportion of tasks.}
  \label{fig:ablation_line}
\end{figure}

\noindent
\textbf{Number of WKT Tags and Tasks. }
To better understand the impact of tags and tasks in the Condor pipeline, 
we conduct ablation experiments by incrementally increasing the proportions of Tasks and WKT Tags. 
As shown in \Cref{fig:ablation_line}, while there are some fluctuations, 
the fitted average performance consistently improves with the increasing number of tags and tasks, indicating that both factors contribute positively to enhancing the model's human-preference capabilities.
Furthermore, the influence of tasks is observed to be more significant than that of tags, as evidenced by the larger performance improvements associated with adding tasks. 
Detailed scores are provided in \Cref{tab:tags,tab:tasks}.

\subsection{Self Iteration}\label{self iter}
We conduct self-iteration experiments using Condor data generated by Qwen2.5-7B and Qwen2.5-72B to evaluate whether a single model can enhance its chat capabilities by itself. 
We train the Base model and then report human-preference performance scores across 8 benchmarks. 
As shown in \Cref{tab:selfiter}, both the 7B and 72B models achieve self-improvement even compared with the RLHF models. 
Notably, the improvement for the 72B model is slightly greater than that of the 7B model, suggesting that larger, more capable models are better equipped to achieve self-improvement.

\begin{table}[!t]
    \centering
    \adjustbox{width=0.5\textwidth}{
        \begin{tabular}{@{\extracolsep{\fill}}lcccc}
            \toprule
            \textbf{Datasets} & \textbf{7B-IT} & \textbf{7B-Condor} & \textbf{72B-IT} & \textbf{72B-Condor} \\ \midrule
        CompassArena & \textbf{33.80} & 32.83 & 43.70 & \textbf{48.73}  \\
        FoFo & 0.52 & 0.45 & \textbf{0.66} & 0.60 \\
        AlignBenchv1.1 & \textbf{6.22} & 6.02 & 6.94 & \textbf{6.98} \\
        AlpacaEvalv2 & 34.66 & \textbf{41.37} & 47.64 & \textbf{56.15}  \\
        ArenaHard & 53.65 & \textbf{60.39} & \textbf{85.67} & 84.86\\
        FollowBench & 0.84 & 0.84 & \textbf{0.92} & 0.91  \\
        MTBench101  & \textbf{8.60} & 8.38 & \textbf{8.45} & 8.44  \\
        WildBench & 16.71 & \textbf{19.78} & 45.23 & \textbf{47.75}    \\
        \rowcolor{tabhighlight}\textbf{Average} & 58.02 & \textbf{58.38} & 70.14 & \textbf{71.12}  \\
        \bottomrule
        \end{tabular}
    }
    \vspace{-2mm}
\caption{\textbf{Condor Self Iteration Experiments on Qwen 7B and 72B.} Here, "IT" refers to the Instruct model after official RLHF. We use GPT-4o as the judge.}
    \vspace{-2mm}
    \label{tab:selfiter}
\end{table}
\section{Analysis and Discussion}



\subsection{Which capability is improved the most?}

 \begin{figure}[t]
  \centering  
\includegraphics[width=0.45\textwidth]{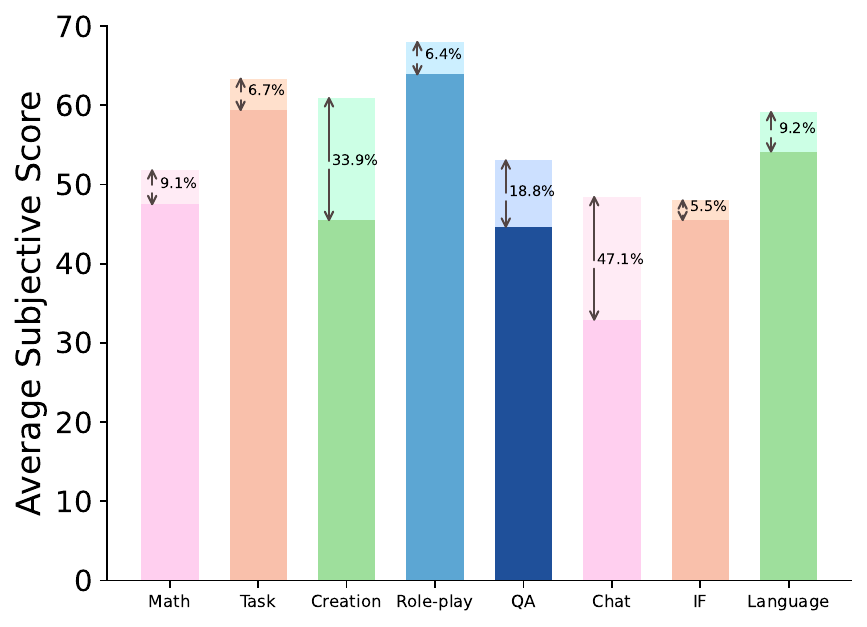}
\vspace{-2mm}
\caption{\textbf{Qwen2.5-7B Performance improvements on Various Sub-Dimensions.} Condor Refine consistently improves the model among all the sub-dimensions. }
  \label{fig:incremental2}
  \vspace{-2mm}
\end{figure}
We evaluate the scores for each benchmark across different sub-capability dimensions and observe improvements in sub-capabilities after Condor training (detailed information in \Cref{subcateappendix}). 
As shown in \Cref{fig:incremental2}, 
the model demonstrates improvement across all sub-capabilities, 
with the most significant score increases observed in the tasks of Creation, QA, and Chat. 
These areas align closely with the core aspects of the model's human-preference capability, highlighting Condor's 
effectiveness in enhancing key conversational skills.


\subsection{How large is the synthetic data coverage?}
\begin{figure}[t]
  \centering  
\includegraphics[width=0.4\textwidth]{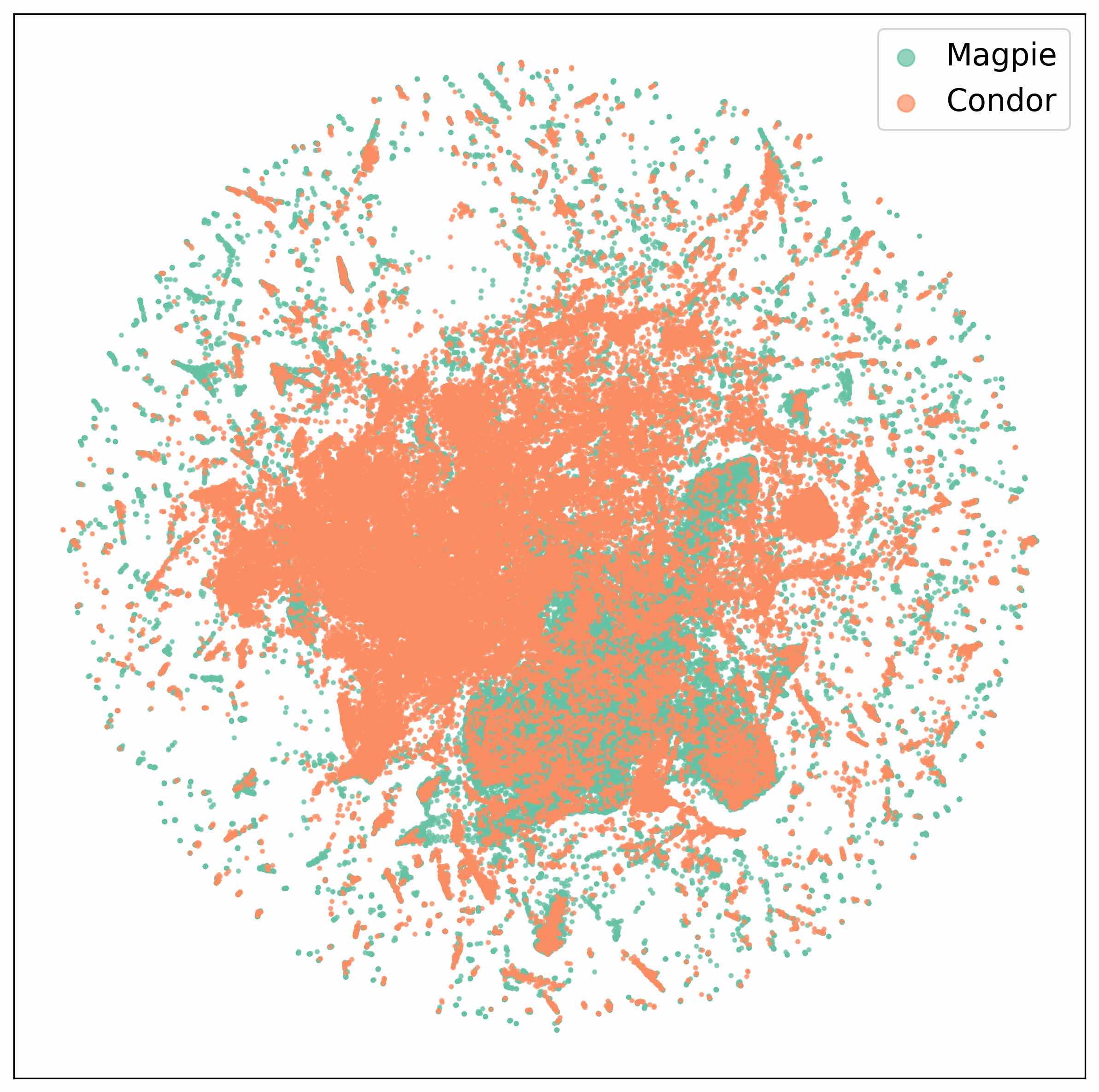}
\caption{\textbf{t-SNE visualization of Condor and Magpie. } Condor shows great alignment with Magpie, highlighting its large diversity and generalization capabilities.}
  \label{fig:tsne}
\end{figure}

We extract the embeddings of questions from Condor and Magpie and use t-SNE for dimensionality reduction and visualization. 
To ensure a fair and balanced comparison, we randomly sample 200k questions from each dataset for this analysis. 
Note that according to Magpie's claimed method, it is capable of feedbacking the model's own training data, which means that the data distribution obtained by Magpie is somewhat close to the model's original SFT data. As shown in \Cref{fig:tsne}, Condor has a similar or even broader distribution compared to Magpie, which indicates that the questions generated by Condor have good diversity.

\subsection{What does SFT actually enhance?}

A comprehensive analysis of the experimental results reveals important insights, particularly from the knowledge-based evaluation results (\Cref{tab:obj ablation}). 
The findings indicate that the SFT phase has minimal impact on the model's intrinsic knowledge capabilities. 
This suggests that the vast majority of knowledge acquisition occurs during the pre-training phase. 
These observations suggest that the SFT phase of Condor contributes very little to enhancing the model's foundational knowledge. Instead, its primary focus is on refining the model’s ability to utilize existing knowledge to answer questions effectively, which can be achieved with a relatively small amount of high-quality data.



\section{Conclusion}

In this paper, we propose Condor, a two-stage data synthesis engine to generate high quality data for supervised fine-tuning. 
Extensive experiments demonstrate that with the high quality data generated by Condor, the performance of the fine-tuned model surpasses many existing methods and the official RLHF models using a small amount of synthetic data. 
We also explore the scaling of synthetic data and self-iteration experiments, demonstrating that models can achieve self-iteration through synthetic data.

\section{Limitations}
Despite the significant improvements brought by Condor in human preference performance, there are still many experiments that require further exploration, such as the use of multi-round iterative synthetic data and how to further enhance the diversity of the synthetic data.
Additionally, the hallucinations produced by LLMs in synthetic data could also become a potential risk. These issues need to be further addressed and improved, which will in turn enhance the quality of the synthetic data.

 \newpage

\bibliography{custom}

\newpage\onecolumn
\appendix
\section{Appendix}

\subsection{Detailed Calculation Method for the Subjective Evaluation Scores.}\label{scoreappendix}
Since the metrics for statistical scoring of various subjective chat datasets are not the same, and the scoring ranges are not all 0-100, we use the following mapping relationships when calculating the average subjective chat score for the model:
\begin{itemize}
    \item The scoring ranges for CompassArena, AlpacaEvalv2, and ArenaHard are 0-100, and no special treatment is needed when calculating the overall mean score.
    \item For FoFo and Followbench, the scoring range is 0-1, and we multiply by 100 when calculating the overall mean score. 
    \item For AlignBenchv1.1 and MTBench101, the scoring range is 0-10, and we multiply by 10 when calculating the overall mean score. 
    \item For WildBench, the scoring range is -100 to 100, and we add 100 to the score and then divide by 2 for mapping.
    \item The final calculated total Average score is the mean score of these datasets after they have been mapped to the 0-100 range.
\end{itemize}

\subsection{Detailed Information for Sub-capabilities Improvement of Condor}\label{subcateappendix}
Based on the subscore for each capability dimension provided by each dataset, we aggregated the model's scores by dimension, thereby obtaining the model's scores on each capability dimension across these subjective test datasets. Specifically, according to the different subscores for various capability dimensions provided by different datasets, we aggregated the scores in the following manner:

\begin{itemize}
    \item Math: The math and reasoning capabilities of model, we aggregate the sub-score from AlignBench, CompassArena.
    \item Task: The task problem svoling capability of model, we aggregate from AlignBench, ArenaHard and MTBench101.
    \item Creation: The ability of the model to create various types of content as required is aggregated from AlignBench, CompassArena, MTBench101, and WildBench.
    \item Role-play: The role-playing capability of the model is aggregated from AlignBench, AlpacaEval, and WildBench.
    \item QA: The knowledge-based question-answering capability of the model is aggregated from AlignBench and CompassArena.
    \item Chat: The daily chat capability of the model is aggregated from AlignBench, AlpacaEval, ArenaHard, MTBench101, and WildBench.
    \item IF: The instruction following capability of the model is aggregated from FoFo and FollowBench.
    \item Language: The language understanding and processing capability of the model is aggregated from AlignBench and CompassArena.
\end{itemize}

\newpage

\begin{table*}[!htbp]
    \centering
    \small
    \resizebox{\textwidth}{!}{
    \begin{tabular}{lccccc}
    \toprule
        {\multirow{2}{*}{Datasets}}& {Qwen2.5-7B-Instruct} & \multicolumn{2}{c}{Qwen2.5-7B-Base} & \multicolumn{2}{c}{Qwen2.5-7B-Instruct} \\
        \cmidrule(lr){2-2} \cmidrule(lr){3-4} \cmidrule(lr){5-6}
         & Official & Condor Void & Condor Refine & Condor Void & Condor Refine \\
    \midrule
        CompassArena & 34.95 & 33.78 & 43.62 & 38.17 & \textbf{46.68} \\ 
        FoFo & 0.45 & 0.38 & 0.48 & \textbf{0.50} & 0.48 \\ 
        AlignBenchv1.1 & 6.16 & 6.08 & 6.26 & \textbf{6.32} & \textbf{6.32} \\
        AlpacaEvalv2 & 32.42 & 33.04 & 60.00 & 43.98 & \textbf{63.85}  \\ 
        ArenaHard & 54.72 & 43.35 & 61.37 & 57.33 & \textbf{63.30} \\
        FollowBench & \textbf{0.86} & \textbf{0.86} & 0.84 & 0.83 & 0.85 \\
        MTBench101 & \textbf{8.37} & 8.12 & 8.18 & 8.28 & 8.24 \\
        WildBench & 15.69 & 13.82 & 31.13 & 18.52 & \textbf{32.99} \\ 
        \rowcolor{tabhighlight} \textbf{Average} & 56.9 & 54.01 & 63.27 & 59.67 & \textbf{64.85} \\

    \bottomrule
    \end{tabular}
    }
    \caption{\textbf{Detailed Results for Condor on Subjective Evaluation Datasets Judged by CompassJudger-1-32B.}}
    \label{tab:main ablation}
\end{table*}

\begin{table*}[htbp]
    \centering
    \small
    \resizebox{0.85\textwidth}{!}{
    \begin{tabular}{lcccccccc}
    \toprule
    \diagbox{Datasets}{Proportion} & 1\% & 2.5\% & 5\% & 10\% & 25\% & 50\% & 100\%  \\
    \midrule
    CompassArena & 20.02 & 40.80 & 43.35 & 41.82 & 40.22 & 42.98 & \textbf{43.62}  \\
    FoFo & 0.18 & 0.41 & 0.42 & 0.47 & 0.46 & \textbf{0.48} & \textbf{0.48}  \\
    AlignBenchv1.1 & 5.28 & 6.14 & 6.07 & 6.15 & 6.20 & \textbf{6.27} & 6.26\\
    AlpacaEvalv2 & 22.36 & 56.65 & 55.53 & 58.88 & \textbf{60.62} & 56.89 & 60.00  \\
    ArenaHard &42.00 & 56.99 & 59.27 & 59.56 & \textbf{63.17} & 61.64 & 61.37 \\
    FollowBench & 0.80 & 0.76 & 0.80 & 0.82 & 0.82 & 0.82 & \textbf{0.84}  \\
    MTBench101 & 6.71 & 8.01 & 8.16 & 8.17 & 8.18 & \textbf{8.22} & 8.18  \\
    WildBench & 0.44 & 29.23 & 28.54 & 29.2 & 28.79 & 30.41 & \textbf{31.13}  \\
    \rowcolor{tabhighlight}\textbf{Average} & 44.07 & 59.74 & 60.79 & 62.09 & 62.50 & 62.68 & \textbf{63.27}  \\
    \bottomrule
    \end{tabular}
    }
\caption{\textbf{Performance Comparison under Different Proportions of Training Data.} Here, 100\% indicates the use of the full dataset, which is approximately 200K synthetic data for training, while the other proportions involve random sampling of the full dataset based on the given percentage.}
    \label{tab:totalnumber}
\end{table*}

\begin{table*}[htbp]
    \centering
    \small
    \resizebox{0.9\textwidth}{!}{
    \begin{tabular}{lcccccccc}
    \toprule

    \diagbox{Datasets}{Proportion}  & 12.50\% & 25\% & 37.50\% & 50\% & 62.50\% & 75\% & 87.50\% & 100\%  \\
    \midrule
    
        CompassArena & 42.57 & 40.98 & 41.87 & 42.25 & 42.88 & 41.75 & 43.5 & \textbf{43.62} \\
        FoFo & 0.45 & 0.47 & 0.49 & 0.48 & \textbf{0.50} & 0.46 & 0.48 & 0.48  \\
        AlignBenchv1.1 & 6.14 & 6.18 & 6.18 & 6.24 & 6.25 & 6.19 & 6.14 & \textbf{6.26} \\
        AlpacaEvalv2 & \textbf{60.12} & 55.65 & 55.03 & 56.89 & 56.77 & 57.76 & 58.14 & 60.00  \\
        ArenaHard &62.19 & \textbf{62.43} & 60.81 & 60.01 & 59.73 & 60.11 & 60.99 & 61.37  \\
        FollowBench & 0.80 & 0.86 & 0.85 & 0.83 & \textbf{0.87} & 0.83 & 0.81 & 0.84  \\
        MTBench101 & 8.18 & \textbf{8.20} & 8.15 & 8.17 & 8.14 & 8.18 & \textbf{8.20} & 8.18   \\
        WildBench & 28.83 & 27.96 & 29.70 & 28.97 & 27.15 & 28.89 & 29.25 & \textbf{31.13}  \\
        \rowcolor{tabhighlight}\textbf{Average} & 62.08 & 62.43 & 62.43 & 62.3 & 62.98 & 62.07 & 62.46 & \textbf{63.27}  \\
    \bottomrule
    \end{tabular}
    }
\caption{\textbf{Performance of the model after training with different numbers of tags in the training data.} The total number of Chinese tags is 4249, and the total number of English tags is 4296. We randomly sampled the Chinese and English tags in proportion and then combined the sampled Chinese and English data for training to ensure a balance of tags between the two languages.}
    \label{tab:tags}
\end{table*}

\begin{table*}[htbp]
    \centering
    \small
    \resizebox{0.85\textwidth}{!}{
    \begin{tabular}{lcccccccc}
    \toprule
    \diagbox{Datasets}{Task Number}   & 1 & 2 & 3 & 4 & 5 & 6 & 7  \\
    \midrule
    
        CompassArena & 42.83 & 42.62 & 41.1 & 42.63 & 42.85 & \textbf{43.88} & 43.62  \\
        FoFo & 0.42 & 0.42 & 0.44 & 0.44 & \textbf{0.49} & 0.46 & 0.48  \\
        AlignBenchv1.1 & 6.18 & \textbf{6.28} & 6.24 & 6.15 & 6.20 & 6.16 & 6.26 \\
        AlpacaEvalv2 & 56.65 & 58.26 & 59.75 & 56.65 & 57.64 & 58.51 & \textbf{60.00}  \\
        ArenaHard &62.6 & 59.40 & 60.86 & 60.61 & 61.60 & \textbf{62.91} & 61.37\\
        FollowBench & 0.81 & 0.83 & 0.76 & 0.75 & 0.78 & 0.82 & \textbf{0.84}   \\
        MTBench101 & 8.11 & \textbf{8.24} & 8.21 & 8.18 & 8.15 & 8.19 & 8.18  \\
        WildBench & 33.07 & \textbf{33.69} & 31.38 & 31.38 & 29.33 & 30.94 & 31.13  \\
        \rowcolor{tabhighlight}\textbf{Average} & 61.7 & 62.16 & 61.46 & 60.89 & 62.04 & 62.81 & \textbf{63.27}  \\
    \bottomrule
    \end{tabular}
    }
    \caption{\textbf{Performance of the model after training with different numbers of tasks in the training data.} All the results in the table were obtained by testing the Qwen2.5-7B-Base model after training on the relevant datasets, with the number of tasks increasing by inclusion. The first task is Role-Playing, the second task is Daily Chat, the third task is Domain Knowledge Q\&A, the fourth task is Given Material Processing, the fifth task is Response Format Control, the sixth task is View, and the seventh task is Creation.}
    \label{tab:tasks}
\end{table*}



\begin{tcolorbox}[
colback=white!10!white,
colframe=black!50!white,
title=Prompt for Question Synthesis,
]

Now we need to create high-quality SFT data for LLM training, so we need you to produce a batch of such data. You only
need to create Questions. I will give you a theme and some examples of SFT data Questions. You need to create three
Questions of different difficulty levels based on this new theme.\\
Your Questions must meet the following requirements:\\
1. You must strictly create only three Questions at a time. These three Questions must be in the domain of \textcolor{red}{[[domain]]}
and the Questions should align with the given theme of \textcolor{red}{[[theme]]}.\\
2. The Questions you create must have context and sufficient information; they should not be abrupt and directly ask the
question.\\
3. Your reply must strictly follow the format below. Your Questions need to be included between [Question Start] and
[Question End], and the difficulty level should be indicated at the beginning, as in the following format:\\

[Easy][Question Start]Question[Question End]

[Medium][Question Start]Question[Question End]

[Hard][Question Start]Question[Question End]

4. Your Questions of different difficulty levels should be distinct and actually reflect the different levels of difficulty.\\
\quad \\
Here are some examples of high-quality SFT data Questions for \textcolor{red}{[[domain]]}:
\textcolor{red}{[example list]}\\
Now it's your turn. Please provide the three Questions of different difficulty levels you created about the theme of
\textcolor{red}{[[theme]]} for \textcolor{red}{[[domain]]}, according to the requirements. Do not be confined by the theme; your Questions only need to
be related to the theme. You can use your rich imagination, but note that you cannot copy the expression from the
examples; you must have your own new expression:
\label{tab:question_generation}
\end{tcolorbox}

\begin{tcolorbox}[
colback=white!10!white,
colframe=black!50!white,
title=Prompt for Response Critic,
breakable]

There is now a user’s question and a model’s response. You need to write a critique for this response, pointing out the
strengths and weaknesses of the model’s answer to help the model improve its response.

Your critique must strictly adhere to the following format:

[Critique Start]

[Strength Start]Strength[Strength End]

[Weakness Start]Weakness[Weakness End]

[Suggestion Start]Suggestion[Suggestion End]

[Critique End]

That is, you must place the strength, weakness, and suggestion in their respective sections.

Here is the user’s question and the model’s response: \textcolor{red}{[dialogue]}

Now it’s your turn. Please provide your Critique as required:
\label{tab:critique_prompt}
\end{tcolorbox}

\clearpage

\begin{tcolorbox}[
colback=white!10!white,
colframe=black!50!white,
title=Condor Void Example - User Question,
breakable]
\label{example}
I was watching a documentary about mapping ancient cities, and it got me thinking about how
maps have evolved over time. Have you ever seen an old map and marveled at how different it looks
from today's maps? What's the oldest map you've ever seen?
\label{fig:patterne}
\end{tcolorbox}

\begin{tcolorbox}[
colback=white!10!white,
colframe=black!50!white,
title=Condor Void Example - Origin Answer,
breakable]

As a digital AI assistant, I don't have personal experiences, emotions, or visual interactions, so I
haven't "seen" maps in the classical sense. I don't have a physical presence, and my interactions
are limited to text-based conversations. However, I'd be delighted to take you on a journey
through the evolution of maps and highlight some notable examples of ancient and historic maps!

$\\$
The Evolution of Maps

1. Prehistory (~20,000 BCE - 3,000 BCE):
Early maps were often simple, symbolic, and focused on specific areas, like the Turin Papyrus Map
(~1160 BCE), which depicts the location of gold mines in ancient Egypt.

2. Ancient Civilizations (~3,000 BCE - 500 CE):
- Babylonian Map of the World (~600 BCE): A clay tablet showing the world as a flat disk,
surrounded by an ocean.
- Ptolemy's World Map (~150 CE): A more accurate, yet still flawed, representation of the
known world, using longitude and latitude.

3. Medieval Period (~500 - 1500 CE):
- Mappa Mundi (~1300 CE): A symbolic, Christian-centric map, often displaying Jerusalem at the
center.

4. Age of Exploration (~1500 - 1700 CE):
- Waldseemüller Map (1507 CE): One of the earliest maps to depict the New World, naming
America after Amerigo Vespucci.

5. Modern Cartography (~1700 CE - present):

- Triangulation and Surveying: Led to more accurate mapping of territories.

- Aerial Photography and Satellite Imaging (20th century): Revolutionized mapmaking with high-
resolution, globally consistent data.

- Digital Maps (late 20th century - present): Enabled interactive, dynamically updated, and
highly accurate maps, like Google Maps.

$\\$
Some of the Oldest Known Maps

- Cave of El Castillo Map (~14,000 BCE): A simple, prehistoric map etched into a cave wall in
Spain, showing the location of a nearby water source.

- The Bedolina Map (~5,000 BCE): A rock carving in Italy, depicting a valley with topographical
features.

- The Nubian Map (~2,000 BCE): An ancient Egyptian map, showing the location of Nubian gold
mines and territories.

$\\$
Marveling at Old Maps

While I may not have personal experiences, I can certainly help you appreciate the historical
context, artistic value, and cartographic innovations of old maps. If you'd like to discuss a specific
map or era in more detail, feel free to ask!

$\\$
Which aspect of old maps fascinates you the most? Is it:

•A) The artistic and symbolic representations

•B) The historical context and cultural influences

•C) The cartographic innovations and technological advancements

•D) Something else (please share!)
\label{fig:patterne}
\end{tcolorbox}

\clearpage


\begin{tcolorbox}[
colback=white!10!white,
colframe=black!50!white,
title=Condor Refine Example - Reflection,
breakable]
[Strength Start]

Strengths:

1.Comprehensive Response:
The model provides an exhaustive overview of the evolution of maps, covering various historical periods, notable examples, and technological
advancements.

2.Structured Formatting:
The use of clear headings, numbered lists, and concise bullet points enhances readability and facilitates easy understanding of complex
information.

3.Engagement Initiator:
The response ends with an interactive element, encouraging user participation and potentially leading to a more in-depth discussion.

4.Honesty About Limitations:
The model transparently acknowledges its limitations as a digital AI assistant, setting clear expectations for the user.

[Strength End]

$\\$

[Weakness Start]

Weaknesses:

\textcolor{red}{1.Length and Information Overload:}
The response is lengthy and packed with numerous examples, which might overwhelm the user. Some points could be elaborated upon in
subsequent interactions rather than all at once.

\textcolor{red}{2.Lack of Visual Enhancement:}
Despite discussing maps, the response is text-only. Incorporating images, diagrams, or even suggesting external visual resources could
significantly enhance the user's understanding and engagement.

\textcolor{red}{3.Initial Acknowledgement Could Be Brief:}
While honesty about the model's limitations is appreciated, the initial acknowledgement could be more concise to quickly transition to the
more engaging and informative sections.

\textcolor{red}{4.Question Response Mismatch:}
The user inquired about the oldest map the model has "seen," which the model addresses by stating its inability to see. However, the model
could more directly address the implicit curiosity about old maps by initially highlighting one or two of the oldest known maps before delving
into the broader evolution.

[Weakness End]

$\\$
[Suggestion Start]

Suggestions for Improvement:

\textcolor{red}{1.Tiered Information Disclosure:}
Initially provide a brief overview of the evolution of maps and highlight 2-3 of the oldest known maps. Offer the user the option to explore
specific eras or types of maps in more detail, facilitating a more interactive and paced information exchange.

\textcolor{red}{2.Incorporate Visual Aids or References:}
Suggest reputable online resources or include descriptions that encourage users to visually explore the mentioned maps, enhancing their
understanding of cartographic developments.

\textcolor{red}{3.Refine the Initial Limitation Disclosure:}
Condense the initial acknowledgement to a single sentence, swiftly moving the focus to the engaging content (e.g., "As a text-based AI, I'll
guide you through the fascinating evolution of maps, highlighting notable examples and innovations.").

[Suggestion End]

\label{fig:patterne}
\end{tcolorbox}

\clearpage
\begin{tcolorbox}[
colback=white!10!white,
colframe=black!50!white,
title=Condor Refine Example - Refined Answer,
breakable]

Discovering the Ancient Roots of Modern Maps

$\\$
\textcolor{red}{As a text-based AI, I'll guide you through the fascinating evolution of maps, highlighting notable examples and innovations.}

$\\$
\textcolor{red}{Let's Start with the Oldest Known Maps:}

1.Cave of El Castillo Map (~14,000 BCE):

Etched into a Spanish cave wall, this simple map shows the location of a nearby water source.

2.The Bedolina Map (~5,000 BCE):

A rock carving in Italy, depicting a valley with topographical features.

$\\$
\textcolor{red}{Visual Exploration Suggestion:} Visit the British Museum's Online Collection or Google Arts \& Culture to see high-resolution images of these
ancient maps.**

$\\$
Curious About the Evolution of Maps?
Here's a concise overview. Feel free to pick any era to explore in more depth:

- Prehistory (~20,000 BCE - 3,000 BCE): Simple, symbolic maps (e.g., Turin Papyrus Map)

- Ancient Civilizations (~3,000 BCE - 500 CE): Notable examples include Babylonian Map of the World and Ptolemy's World Map

- Medieval to Modern (~500 - present): Advances from Mappa Mundi to Triangulation, Aerial Photography, and Digital Maps like Google
Maps

$\\$
Which Era or Aspect of Old Maps Would You Like to Explore Further?

•A) Ancient symbolic representations

•B) Medieval navigational charts

•C) The leap to digital cartography

•D) Something specific (please share, and I'll delve deeper!)

\label{fig:patterne}
\end{tcolorbox}

\clearpage


\begin{tcolorbox}[
colback=white!10!white,
colframe=black!50!white,
title=Detailed explanation of total 7 chat tasks,
breakable]
\label{taskdef}
[Task 1-Role Playing]

Engage in simulated conversations or behaviors by assuming different roles to
explore various perspectives or scenarios.

$\\$
[Task 2-Daily Chat]

Participate in casual conversations, including greetings, small talk, and sharing
everyday experiences.

$\\$
[Task 3-Domain QA]

Provide specialized and accurate answers to questions within a specific domain or
field.

$\\$
[Task 4-Given Material Processing]

Analyze, process, and summarize given texts, data, or other materials.

$\\$
[Task 5-Response Format Control]

Adjust the format, style, and expression of responses according to specified
requirements.

$\\$
[Task 6-View]

Offer personal opinions, insights, or perspectives on a particular topic.

$\\$
[Task 7-Creation]

Generate new content such as articles, stories, poetry, designs, etc.

\label{fig:patterne}
\end{tcolorbox}


\clearpage

\begin{tcolorbox}[
colback=white!10!white,
colframe=black!50!white,
title=Comparison Example - User Question,
breakable]

Is the ATF a violation in of itself against the American people?

\label{fig:patterne}
\end{tcolorbox}

\begin{tcolorbox}[
colback=white!10!white,
colframe=black!50!white,
title= Comparison Example - Qwen2.5-7B-Instruct Response,
breakable]

The term "ATF" typically refers to the Bureau of Alcohol, Tobacco, Firearms and Explosives, a
federal agency within the United States Department of Justice. The ATF is responsible for
reducing crime and preventing violence by reducing the illegal flow and improper use of
firearms, alcohol, tobacco, and explosives.

$\\$
The ATF itself is not a violation against the American people. Rather, it is an agency that
operates under the laws and regulations established by the U.S. government to enforce laws
related to alcohol, tobacco, firearms, and explosives. The agency's actions are generally aimed
at preventing crime and ensuring public safety.

$\\$
However, like any government agency, the ATF can be subject to criticism and controversy.
Some criticisms have been raised regarding the agency's enforcement practices, the impact
of its regulations on law-abiding citizens, and concerns about civil liberties. These criticisms
do not imply that the agency itself is a violation, but rather that its actions or policies may be
subject to scrutiny and debate.

$\\$
If you have specific concerns or questions about the ATF's actions or policies, it would be
helpful to provide more context or details so that a more accurate and informed response can
be given.
\label{fig:patterne}
\end{tcolorbox}

\begin{tcolorbox}[
colback=white!10!white,
colframe=black!50!white,
title=Comparison Example - Qwen2.5-7B-Condor-Refine Response,
breakable]

The term "ATF" typically refers to the Bureau of Alcohol, Tobacco, Firearms and Explosives, a
federal agency in the United States responsible for enforcing federal laws related to the
manufacture, possession, and use of alcohol, tobacco, firearms, and explosives. The ATF is not
a violation against the American people; rather, it is an agency that operates within the legal
framework established by the U.S. Congress to enforce specific laws.

$\\$
However, the ATF's operations and the laws it enforces can be subjects of debate and
criticism. For example, some argue that the agency's enforcement of gun control laws
infringes on the Second Amendment rights of law-abiding citizens. Others may criticize the
agency for its handling of certain cases or its budget and resource allocation. Additionally,
there have been instances where the ATF has been involved in controversial operations, such
as the Ruby Ridge and Waco incidents, which have fueled public debate and criticism.

$\\$
\textcolor{red}{It's important to note that the ATF's role is to enforce the laws as they are written, and the
laws themselves are subject to interpretation and debate.} The agency's actions are generally
aimed at reducing crime and ensuring public safety, but the effectiveness and fairness of
these actions are often subjects of public and political discourse.

$\\$
\textcolor{red}{If you have specific concerns or questions about the ATF's operations or the laws it enforces,
it might be helpful to look at the broader context of U.S. law and policy, including the
legislative history and the ongoing debates surrounding gun control and public safety. For more
detailed information, you can refer to official ATF publications, such as the ATF's annual
reports, or consult reputable news sources and legal analyses."}

\label{fig:patterne}
\end{tcolorbox}





\end{document}